\documentclass[]{llncs}

\usepackage[T1]{fontenc}
\usepackage{tabularx}
\usepackage{array}
\usepackage{xcolor}
\usepackage{listings}
\usepackage{amsmath,amsfonts,amssymb}
\usepackage{graphicx, tcolorbox}
\usepackage{booktabs}
\usepackage{epstopdf}
\usepackage{enumitem}
\usepackage{algorithm}
\usepackage{algorithmic, amsmath}
\usepackage{colortbl}
\usepackage{multirow}
\usepackage[table]{xcolor} 
\usepackage{threeparttable}
\usepackage{adjustbox}
\definecolor{bestblue}{RGB}{0,92,175}
\definecolor{lightgray}{RGB}{245,245,245}
\renewcommand{\footnotesize}{\fontsize{8.5pt}{9pt}\selectfont}

\lstdefinestyle{rules}{
  basicstyle=\ttfamily\footnotesize,
  columns=fullflexible,
  keepspaces=true,
  frame=single,
  framerule=0.3pt,
  rulecolor=\color{black!30},
  backgroundcolor=\color{black!2},
  xleftmargin=0.6em,
  xrightmargin=0.6em,
  aboveskip=0.5em,
  belowskip=0.5em,
  showstringspaces=false,
  breaklines=true
}

\begin{document}
\title{Learning Probabilistic Logic Programs with Functional Gradient Guided Language Models}
%
\author{Saurabh Mathur$^{*}$ \inst{1} \and Sahil Sidheekh$^{*,\dagger}$ \inst{2,3} \and  Bhavan Vasu$^{*}$ \inst{4} \and Farbod Tavakkoli\inst{3} \and Prasad Tadepalli \inst{4} \and Kristian Kersting \inst{1} \and Sriraam Natarajan \inst{2}}
\institute{ Technische Universität Darmstadt \and The University of Texas at Dallas \and AT\&T CDO \and  Oregon state University}


\maketitle              

\begingroup
\renewcommand\thefootnote{\fnsymbol{footnote}} 
\footnotetext[0]{$^*$ These authors contributed equally to this work.}
\footnotetext[0]{$^\dagger$ Work done in part while Sahil was an intern at AT\&T CDO.}
\endgroup

\begin{abstract}
Declarative logic programs offer a powerful and interpretable abstraction for encoding relational structure and neurosymbolic reasoning, by expressing dependencies as weighted compositional rules.
However, inducing them from data remains fundamentally hard, bottlenecked by the combinatorial explosion of symbolic search spaces. LLMs have recently emerged as powerful hypothesis generators, but when used in isolation, they lack the capacity to do systematic inductive reasoning needed to reliably synthesize valid programs that fit complex relational distributions. We introduce \textsc{grasp} (Gradient-boosted Synthesis of Probabilistic
logic programs), a neurosymbolic framework that casts relational
structure learning as functional gradient boosting in which the weak
learner is a first-order rule and the intractable inner search is
delegated to an LLM proposal oracle.
We evaluate \textsc{grasp} on four relational benchmarks spanning molecular toxicity prediction (Tox21), mutagenesis, and citation matching (Cora), and show that it improves over purely symbolic, neural, and LLM-based baselines, 
while producing interpretable weighted rule ensembles. By replacing combinatorial search with gradient-guided LLM hypothesis generation, \textsc{grasp} retains boosting guarantees without sacrificing the transparency of symbolic outputs.
\end{abstract}
\section{Introduction}
Relational data arises naturally in many scientific domains, including biological interaction networks, citation graphs, and pharmacological databases. Probabilistic logic programs are a natural representation for such data, expressing relational dependencies as interpretable, compositional rules with calibrated confidences. The predicates of such programs can be grounded in diverse data sources and modalities, from structured relational databases to unstructured inputs abstracted into deep neural predicates~\cite{manhaeve2021}. However, inducing these programs from data remains a fundamental challenge~\cite{marra2024}.

The key difficulty is that of \emph{structure learning}: determining which clauses constitute the program. Traditional inductive logic programming (ILP) and statistical relational learning (SRL) methods address this challenge by searching over a hypothesis space of candidate rules. However, this space grows exponentially with rule length, predicate arity, and relational depth, making exhaustive search intractable for complex domains \cite{cropper2022inductive}. Functional gradient boosting \cite{friedman2001greedy,natarajan2012} has offered a principled decomposition of this problem, reformulating structure learning as a sequence of regression problems, each requiring a symbolic weak learner to propose the next rule~\cite{natarajan2012}. While this decomposition reduces the scope of search at each step, 
inducing the weak learner still requires symbolic search over the relational hypothesis space, which remains the dominant computational bottleneck as relational complexity grows.

Pretrained large language models (LLMs) have emerged as effective hypothesis generators owing to their broad semantic knowledge from their vast training corpora ~\cite{qiu2024phenomenal,luo2025}. However, when used in isolation, LLMs lack a mechanism to ground rule generation in the training distribution. When prompted directly, they can produce semantically plausible but inaccurate rules~\cite{helff2025}, with no guarantees that a generated clause is valid, executable, or consistent with observed data.  This tension, between powerful but ungrounded proposers on one hand and principled but search-bound learners on the other, motivates our approach.

We introduce \textsc{grasp} (Gradient-boosted Synthesis of Probabilistic Logic Programs with Language Models), a neurosymbolic framework that addresses this gap by casting relational structure learning as functional gradient boosting in which the weak learner is a first-order rule and the intractable inner combinatorial search is achieved via LLM proposals.
At each boosting iteration, \textsc{Grasp} computes pseudo-residuals over the current program ensemble, identifies the examples carrying the largest functional gradient, and conditions the LLM on local relational descriptions of those examples to propose a small candidate set of rules. Each candidate is symbolically evaluated against the data to learn their weights, and admitted to the ensemble only if it reduces the residual loss. The LLM thus serves purely as a proposal distribution while the boosting framework with symbolic verification remains the trusted estimator. This equips the framework with convergence guarantees of functional gradient boosting, while overcoming the hallucination challenges with purely LLM-based rule generation.
Our key contributions are as follows:

\begin{itemize}[leftmargin=*, itemsep=2pt]

\item We introduce \textsc{grasp}, a framework that uses functional gradient boosting to steer LLM-driven synthesis of probabilistic logic programs for relational structure learning.

\item We show that pseudo-residual-guided example selection provides an effective signal for directing LLM rule proposal toward unexplained relational variance, replacing combinatorial symbolic search at each boosting step.


\item We evaluate \textsc{grasp} on $3$ relational datasets spanning molecular toxicity prediction (Tox21), mutagenesis, and citation matching (Cora), outperforming symbolic, neural, and LLM-based baselines while producing interpretable weighted rule ensembles.

\end{itemize}



\begin{figure*}[t]
    \centering
    \includegraphics[width=\linewidth]{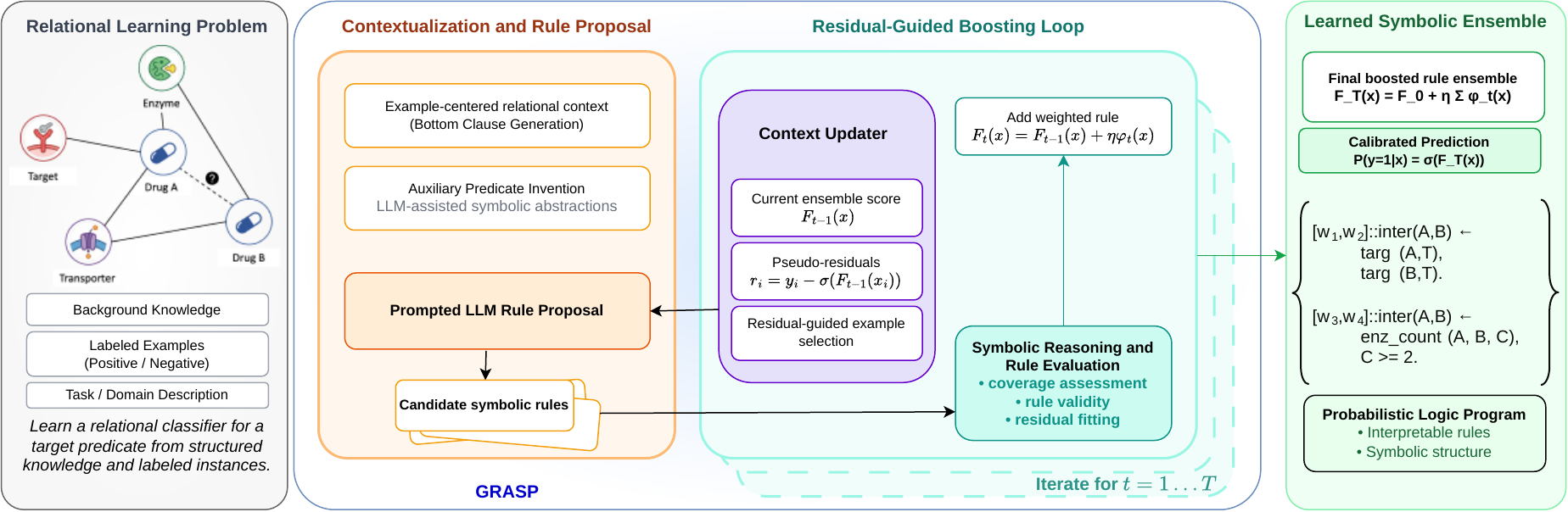}
    \caption{{\bf \textsc{grasp} Framework.}
     \textsc{grasp} frames relational learning as a neurosymbolic functional gradient boosting problem. At each iteration $t$, an LLM receives an example-centered relational context and proposes candidate symbolic rules. A logic solver evaluates each candidate against the relational data, assessing coverage and residual fit. The context updater computes pseudo-residuals $r_i = y_i - \sigma(F_{t-1}(x_i))$ and selects high-residual examples to steer the LLM toward patterns not yet explained by the ensemble. The accepted rule $\varphi_t$ is added with step size $\eta$, yielding $F_t(x) = F_{t-1}(x) + \eta\varphi_t(x)$. After $T$ iterations, \textsc{grasp} returns a weighted ensemble of interpretable probabilistic logic rules. This design replaces combinatorial symbolic search with gradient-guided LLM hypothesis generation, while retaining boosting convergence guarantees.
    }
    \label{fig:GRASP}
\end{figure*}
\section{Background and Related Work}
\subsection{Statistical Relational Learning}
Inductive logic programming addresses the problem of inducing definite logic programs from relational examples, searching over a hypothesis space of candidate clauses consistent with background knowledge and labeled instances~\cite{muggleton1991,srinivasan2001aleph}. Classical ILP operates in a noise-free setting, that is, a clause is acceptable if it covers all positive examples and no negative examples. This assumption is brittle in practice, where data is noisy, observations are incomplete, and the target relation may not be deterministically entailed by any finite set of clauses.

Statistical relational learning extends this setting by combining first-order logic with probabilistic reasoning, explicitly modeling uncertainty over relational data \cite{de2016statistical,marra2024}. We work with a first-order language $\mathcal{L}$ defined over predicate symbols $\mathcal{P}$, constant symbols $\mathcal{C}$, and variable symbols $\mathcal{V}$. A \emph{probabilistic logic program} is a finite set of definite clauses $H \leftarrow B_1, \ldots, B_m$, each associated with a real-valued weight $w \in \mathbb{R}$. Given a set of ground atoms, predicted probabilities are obtained via a sigmoid transformation over the weighted sum of clause activations, allowing the model to assign graded confidence to predictions rather than hard entailment. An SRL learning task is typically specified by background knowledge $B$, positive examples $\mathcal{E}^+$, negative examples $\mathcal{E}^-$, and a set of labeled examples $\mathcal{E} = {(e_i, y_i)}_{i=1}^{n}$ with $y_i \in {0,1}$.

Given a fixed program structure, \emph{parameter learning} estimates clause weights by optimizing a differentiable loss over the labeled examples, such as the negative conditional log likelihood. This is a tractable continuous optimization problem, given the clauses and their groundings. While these clauses can be specified from domain knowledge in some domains, obtaining them remains challenging in less well-understood domains. \emph{Structure learning} induces the clause structure from data. This requires searching over a hypothesis space of candidate clauses that grows exponentially with rule length, predicate arity, and relational depth.
Functional gradient boosting provides a principled decomposition of structure learning that avoids global search \cite{friedman2001greedy,natarajan2012,khot2011learning}. Rather than inducing a complete program at once, boosting constructs the program iteratively: at each step $t$, pseudo-residuals $r_i^{(t)} = y_i - p_i^{(t-1)}$ are computed over the current ensemble, identifying examples that remain poorly explained. A weak learner $\varphi_t$ is then fit to approximate these residuals, and the ensemble is updated as $F_t = F_{t-1} + \eta\varphi_t$. In statistical relational learning, the weak learner is a relational regression tree whose induction requires symbolic search over the relational hypothesis space \cite{natarajan2012}. This search remains a computational bottleneck as relational complexity grows, and motivates the approach taken by \textsc{grasp}.


\subsection{Neurosymbolic Learning}
Neurosymbolic learning seeks to combine the generalization capabilities of neural networks with the structured reasoning of symbolic systems, addressing the complementary weaknesses of each paradigm in isolation. A key mechanism for this integration is the neural predicate \cite{manhaeve2021}, which allows subsymbolic inputs such as images, text, or molecular graphs to be processed by neural networks whose outputs annotate logical facts with learned probabilities \cite{shindo2023alpha}. While these advances make probabilistic logic programs significantly more expressive, they do not address the problem of inducing program structure from data. The clause structure must still be specified or learned, and doing so from noisy, uncertain, and incomplete relational data remains challenging.

A parallel line of work has sought to overcome the limitations of symbolic search by relaxing discrete rule learning into differentiable optimization \cite{yang2017,sadeghian2019drum,xu2022ruleformer}. In particular, Neural LP \cite{yang2017} uses an end-to-end differentiable framework for learning logical rules over knowledge bases, encoding inference as a sequence of differentiable matrix operations over relation embeddings. This approach avoids explicit symbolic search but remains constrained by a predefined set of rule templates, limiting the expressivity of the hypothesis class to structures enumerated before training begins. Further, Neural LP is restricted to binary predicates; higher-arity relations must be decomposed into equivalent binary predicates by introducing composite entities, an encoding that obscures relational structure and does not scale naturally to complex multi-argument relations. 

Large language models have more recently been proposed as hypothesis generators for rule learning, offering a path to rule synthesis that bypasses exhaustive enumeration entirely \cite{qiu2024phenomenal}. ChatRule \cite{luo2025} exploits the relational and semantic knowledge encoded in LLMs to propose and rank logical rules over knowledge graphs without enumerating a predefined search space. However, this approach relies on one-shot generation with post-hoc ranking and provides no mechanism for iterative feedback grounded in the training distribution. LLMs lack the capacity for systematic inductive reasoning~\cite{helff2025}, and without grounding in the training distribution, they produce rules that are syntactically valid but statistically uninformative. \textsc{grasp} addresses this gap by coupling LLM hypothesis generation with functional gradient boosting, providing the data-grounded feedback signal that prior LLM-based approaches lack.

\subsection{Explainability}
While a rule-based machine learning model can be considered intrinsically interpretable, the explainability of rules from such an induction is not guaranteed. An explainable rule base must satisfy several structural and functional criteria \cite{belle2021principles} to be considered explainable. An ideal explainable rules set must be succinct and minimal, as interpretability relies heavily on cognitive simplicity and degrades rapidly as the size of the rule base or the complexity of individual clauses grows~\cite{margot2021new}. Explainability is also heavily dependent on coverage; rather than generating hyper-specific clauses that overfit to isolated instances, explainable rules capture broad patterns that generalize well across the target distribution. For rules generated in probabilistic frameworks, the interpretability of weights should provide a meaningful rationale that a human can reason about, while the magnitude and sign of such weights should transparently map to the confidence or probability of the target relation. Finally, for a rule to be interpretable and actionable, a human user should be able to mentally trace the logical conditions and weights to verify the model's output without relying on external computation (simulatability), and each predicate, variable, and parameter within the rule should independently represent a coherent, real-world concept (decomposability). \textsc{grasp} satisfies all these components of explainability by incorporating succinctness, coverage, simulatability, and decomposability.
\section{\textsc{grasp}: Gradient-Boosted Synthesis of Probabilistic Logic Programs}



Relational learning tasks are characterized by structured background knowledge, labeled examples over a target predicate, and the need to induce predictive models that generalize over unseen entity combinations. Unlike propositional learning, the hypothesis space is defined over first-order clauses whose complexity grows with the number of predicates, argument types, and relational depth, making exhaustive search intractable.

\begin{tcolorbox}[colback=gray!5, colframe=gray!50, title=\textbf{Problem: Relational Learning}]
\textbf{Given:}
\begin{itemize}[leftmargin=*,itemsep=2pt]
\item A domain description $\mathcal{D}$, a set of entities $\mathcal{C}$, a set of predicate definitions $\mathcal{P},$ and a target predicate $q \in \mathcal{P}$
  \item Structured background knowledge $B$, and a set of labeled examples $\mathcal{E} = {(e_i, y_i)}_{i=1}^{n}$ with $y_i \in {0,1}$
  \item An LLM proposal oracle $\mathcal{O}_{\mathrm{LLM}}$
\end{itemize}
\textbf{To do:} Learn an accurate yet interpretable model structured as the additive ensemble
\begin{equation}\label{eq:ensemble}
  F_T(e) = F_0 + \sum_{t=1}^{T} \eta\, \varphi_t(e)
\end{equation}
where each $\varphi_t$
is a scoring function induced by a first-order clause $h_t$, such that the empirical log-loss $\mathcal{L}(F_T)$ over $\mathcal{E}$
is minimized.
\end{tcolorbox}

\textsc{grasp} addresses this problem via an LLM-guided functional gradient boosting loop, where at each iteration $t$, a new clause $h_t$ is proposed by the LLM, evaluated against the current pseudo-residuals, and added to the ensemble with a learned weight. 

\subsection{Relational Abstraction and Aggregation}
The quality of LLM proposals depends on two factors. First, the LLM needs a predicate vocabulary that is expressive enough to describe meaningful relational patterns, rather than raw low-level predicates that describe individual facts and cannot directly capture the higher-order relational structure that governs the target predicate. Second, the functional gradient must be communicated to the LLM in a form that is both meaningful and compact enough to fit within its context window, since pseudo-residuals are scalar values that indicate where the current ensemble fails but carry no semantic information about the relational patterns that could explain those failures. \textsc{grasp} addresses both requirements through a one-time abstraction phase that enriches the background knowledge with derived aggregate predicates and constructs a clause template for generating local relational descriptions of entities to serve as a compact, semantically grounded representation of the functional gradient at each boosting step.

\textbf{Aggregate Rule Generation.} Raw background predicates typically encode individual facts (for example, a single atom or bond in a molecule) and are too fine-grained for the LLM to compose into discriminative rules within a bounded clause length. \textsc{grasp} therefore first prompts $\mathcal{O}_{\mathrm{LLM}}$ with the predicate schema $\mathcal{P}$, the entity set $\mathcal{C}$, and a natural language description of the domain $\mathcal{D}$ to generate rules/programs that compute meaningful aggregates over entities appearing in the target predicate $q$. These aggregates capture higher-order relational structure that is not directly observable from the base predicates alone, including counts of shared intermediate entities, existence of relational attributes, and cardinalities of relational neighborhoods. Formally, the LLM returns a set of derived predicate definitions $\mathcal{P}^+ = \mathcal{O}_{\mathrm{LLM}}(\mathcal{P}, \mathcal{C}, \mathcal{D})$, and the background knowledge is augmented as $B \leftarrow B \cup \mathcal{P}^+$, after the derived predicate is validated for syntactic well-formedness and runtime executability against $B$.

\textbf{Clause Template Construction.} After generating the aggregate predicate library, \textsc{grasp} prompts  $\mathcal{O}_{\mathrm{LLM}}$ a second time with $\mathcal{P}$, $\mathcal{D}$, and the derived predicates $\mathcal{P}^+$ to produce a clause template $\mathcal{T}$. The clause template specifies which predicates may appear in clause bodies, how their arguments are typed, and which argument combinations are semantically admissible. By incorporating the aggregate predicates generated in Stage 1, $\mathcal{T}$ enables the construction of compact local relational descriptions of entities at each boosting step. Constraining rule proposals to $\mathcal{T}$ ensures that candidate clauses are both syntactically valid and semantically grounded in the domain, substantially reducing the effective hypothesis space without exhaustive enumeration.

\subsection{Relational Context Construction}
At each boosting iteration $t$, \textsc{grasp} constructs a local relational description for a small set of training examples to serve as a symbolic, context-window-compatible representation of the functional gradient. The pseudo-residuals $r_i^{(t)} = y_i - \sigma(F_{t-1}(e_i))$ identify examples where the current ensemble is most uncertain or incorrect. \textsc{grasp} selects the $k$ examples with the highest residual magnitude $|r_i^{(t)}|$, which correspond to the instances whose prediction error would benefit most from a new rule.

For each selected example $e_i$, \textsc{grasp} uses the clause template $\mathcal{T}$ to generate a local relational description: a compact symbolic summary of the relational neighborhood of $e_i$ within the background knowledge $B$, expressed using the predicate vocabulary in $\mathcal{P} \cup \mathcal{P}^+$. These descriptions\footnote{We implement these  descriptions as bottom clauses under restricted mode declarations.} encode the relational context relevant to the target predicate in a form that is both semantically meaningful and compact enough to fit within the LLM's context window. Together, the selected examples and their local relational descriptions constitute the functional gradient signal presented to the LLM at iteration $t$.
 
\subsection{Rule Proposal and Evaluation}
The LLM is prompted with the predicate schema $\mathcal{P} \cup \mathcal{P}^+$, the clause template $\mathcal{T}$, the domain description $\mathcal{D}$, the current ensemble $F_{t-1}$, and the local relational descriptions of the $k$ high-residual examples. The residual values $r_i^{(t)}$ are presented alongside each example description, providing a signed signal that indicates whether the ensemble currently under-predicts or over-predicts each example. The LLM proposes a set of candidate clauses $\mathcal{C}_t.$\footnote{Syntactically invalid responses are discarded.}
Each such rule $h$ partitions the training examples into two groups: those it covers and those it does not, and assigns a separate prediction to each group.  This yields a two-leaf scoring function, which we formalize as follows:
 
\begin{definition}[Rule-Based Scoring Function]
\label{def:scoring-fn}
For a rule $h$ with coverage vector
$\mathbf{m}_h \in \{0,1\}^n$, the associated scoring function
$\varphi_h : \mathcal{E} \to \mathbb{R}$ is:
\begin{equation}\label{eq:phi}
  \varphi_h(e_i;\, w^1, w^0)
   = 
  w^1 \cdot h(e_i)  +  w^0 \cdot \bigl(1 - h(e_i)\bigr)
\end{equation}
where $w^1 \in \mathbb{R}$ is the leaf weight for covered examples
($h(e_i) = 1$) and $w^0 \in \mathbb{R}$ is the leaf weight for
uncovered examples ($h(e_i) = 0$).
\end{definition}

Note that every rule contributes a prediction for \emph{every} example: $w^1$ for those it covers and $w^0$ for those it does not. This differs from standard ILP, where a rule applies only to the examples it covers. The two-leaf structure is essential for the gradient boosting framework, as it allows the ensemble to adjust its predictions on all examples at each iteration. Given a proposed rule $h$ and pseudo-residuals $\mathbf{r} = (r_1, \ldots, r_n)$, the optimal leaf weights that minimize the squared residual error are obtained in closed form as the mean residual over covered and uncovered examples, respectively. The weighted proposal that minimizes the squared residual error is selected as $\varphi_t$, and the ensemble is updated as $F_t(e) = F_{t-1}(e) + \eta \varphi_t(e)$. Figure~\ref{fig:GRASP} summarizes the full \textsc{grasp} framework, illustrating the initialization phase, the residual-guided boosting loop, and the learned symbolic ensemble produced after $T$ iterations.

\section{Empirical Evaluation}

\definecolor{grasphl}{RGB}{225,240,255}
\newcommand{\grasp}{\textsc{Grasp}}
\newcommand{\graspGem}{\grasp{}-Gem}
\newcommand{\graspCdx}{\grasp{}-Cdx}
\definecolor{bestblue}{RGB}{10,26,253} \newcommand{\Best}[1]{\textcolor{bestblue}{\textbf{#1}}} \newcommand{\NA}{--}

\begin{table}[t]
\centering
\caption{Predictive performance in terms of AUC-ROC across citation, mutagenesis, and Tox21 domains. Results are reported as percentages, with mean$\pm$std over three trials. The best completed result for each dataset is shown in blue. Overall, \textsc{GRASP} variants are competitive with or outperform classical relational baselines across most domains, while the one-shot LLM baseline is less stable. ``--'' denotes no usable result.}
\label{tab:aucroc-all-domains}
\footnotesize
\setlength{\tabcolsep}{3.5pt}
\renewcommand{\arraystretch}{1.08}
\resizebox{\textwidth}{!}{%
\begin{tabular}{lccccc}
\toprule
\textbf{Dataset}
& \textsc{BoostedRDN}
& \textsc{NeuralLP}
& \grasp{}-Gem
& \grasp{}-Cdx
& \textsc{One-shot} \\
\midrule
\multicolumn{6}{l}{\textbf{Citation and Mutagenesis domains}} \\
\midrule
\texttt{cora}
& $95.02{\pm}1.96$
& $91.86{\pm}0.00$
& $92.99{\pm}0.51$
& \Best{$97.67{\pm}0.26$}
& $74.78{\pm}21.57$ \\




\texttt{mutagen}
& $82.92{\pm}2.37$
& $79.08{\pm}2.44$
& $88.36{\pm}1.72$
& \Best{$88.41{\pm}1.82$}
& $82.72{\pm}2.85$ \\
\midrule
\multicolumn{6}{l}{\textbf{Tox21 domains}} \\
\midrule
\texttt{nr\_ahr}
& \Best{$81.98{\pm}0.47$}
& $68.80{\pm}0.07$
& $76.11{\pm}1.45$
& $81.41{\pm}0.65$
& $63.81{\pm}7.73$ \\

\texttt{nr\_ar}
& $75.84{\pm}1.36$
& \Best{$78.82{\pm}0.19$}
& $74.81{\pm}1.71$
& $78.11{\pm}3.04$
& $61.73{\pm}10.80$ \\

\texttt{nr\_ar\_lbd}
& $76.46{\pm}3.22$
& $75.82{\pm}0.29$
& \Best{$78.23{\pm}1.76$}
& $77.17{\pm}2.54$
& \NA \\

\texttt{nr\_aromatase}
& \Best{$73.18{\pm}3.56$}
& $71.17{\pm}0.93$
& $72.96{\pm}0.15$
& $73.10{\pm}1.52$
& $59.14{\pm}3.02$ \\

\texttt{nr\_er}
& $65.93{\pm}0.78$
& $64.67{\pm}0.11$
& $61.20{\pm}2.39$
& \Best{$67.52{\pm}1.34$}
& $51.37{\pm}0.37$ \\

\texttt{nr\_er\_lbd}
& $73.26{\pm}1.96$
& $62.01{\pm}1.10$
& $70.69{\pm}0.65$
& \Best{$73.70{\pm}1.44$}
& $53.75{\pm}3.15$ \\

\texttt{nr\_ppar\_gamma}
& $71.18{\pm}5.03$
& $58.54{\pm}0.50$
& $75.25{\pm}6.73$
& \Best{$75.69{\pm}0.78$}
& $54.76{\pm}0.99$ \\

\texttt{sr\_are}
& \Best{$68.73{\pm}2.07$}
& $65.06{\pm}0.06$
& $67.31{\pm}1.99$
& $65.70{\pm}2.41$
& \NA \\

\texttt{sr\_atad5}
& $71.41{\pm}1.02$
& $65.53{\pm}0.74$
& $65.77{\pm}3.73$
& \Best{$74.79{\pm}0.21$}
& $55.34{\pm}7.19$ \\

\texttt{sr\_hse}
& $73.83{\pm}1.05$
& $59.26{\pm}0.20$
& $68.38{\pm}2.32$
& \Best{$74.41{\pm}2.93$}
& $56.70{\pm}5.07$ \\

\texttt{sr\_mmp}
& $76.43{\pm}0.90$
& $69.10{\pm}0.06$
& $73.62{\pm}5.97$
& \Best{$79.32{\pm}0.70$}
& $63.16{\pm}0.55$ \\

\texttt{sr\_p53}
& $72.83{\pm}2.50$
& $70.35{\pm}0.21$
& $71.50{\pm}5.46$
& \Best{$74.94{\pm}1.24$}
& $63.98{\pm}3.00$ \\
\bottomrule
\end{tabular}%
}
\end{table}

\begin{table}[t]
\centering
\caption{Predictive performance in terms of AUC-PR across citation, mutagenesis, and Tox21 domains. Results are reported as percentages, with mean$\pm$std over three trials. Since AUC-PR is more sensitive to class imbalance, these results highlight the ability of each method to recover positive examples in sparse relational prediction settings. The best result for each dataset is shown in blue bold. ``--'' denotes no usable result.}
\label{tab:aucpr-all-domains}
\footnotesize
\setlength{\tabcolsep}{3.5pt}
\renewcommand{\arraystretch}{1.08}
\resizebox{\textwidth}{!}{%
\begin{tabular}{lccccc}
\toprule
\textbf{Dataset}
& \textsc{BoostedRDN}
& \textsc{NeuralLP}
& \grasp{}-Gem
& \grasp{}-Cdx
& \textsc{One-shot} \\
\midrule
\multicolumn{6}{l}{\textbf{Citation and Mutagenesis domains}} \\
\midrule
\texttt{cora}
& $96.10{\pm}1.57$
& $94.15{\pm}0.00$
& $94.36{\pm}0.37$
& \Best{$98.23{\pm}0.00$}
& $78.89{\pm}17.77$ \\




\texttt{mutagen}
& $86.53{\pm}1.47$
& $89.53{\pm}1.48$
& \Best{$94.36{\pm}0.65$}
& $93.65{\pm}0.77$
& $90.23{\pm}2.51$ \\
\midrule
\multicolumn{6}{l}{\textbf{Tox21 domains}} \\
\midrule
\texttt{nr\_ahr}
& \Best{$38.17{\pm}1.57$}
& $18.43{\pm}0.35$
& $32.37{\pm}0.72$
& $34.49{\pm}1.02$
& $21.32{\pm}5.72$ \\

\texttt{nr\_ar}
& \Best{$41.99{\pm}2.41$}
& $13.32{\pm}0.02$
& $38.77{\pm}3.02$
& \Best{$41.99{\pm}1.93$}
& $13.79{\pm}11.05$ \\

\texttt{nr\_ar\_lbd}
& $38.74{\pm}1.62$
& $11.54{\pm}0.23$
& \Best{$44.90{\pm}1.96$}
& $40.86{\pm}5.53$
& \NA \\

\texttt{nr\_aromatase}
& $17.67{\pm}1.33$
& $12.35{\pm}0.08$
& \Best{$21.45{\pm}2.49$}
& $18.36{\pm}0.95$
& $10.70{\pm}2.50$ \\

\texttt{nr\_er}
& $24.32{\pm}0.97$
& $19.82{\pm}0.10$
& $23.84{\pm}0.46$
& \Best{$28.97{\pm}3.13$}
& $14.62{\pm}1.16$ \\

\texttt{nr\_er\_lbd}
& $18.61{\pm}1.19$
& $7.32{\pm}0.35$
& $21.27{\pm}3.73$
& \Best{$24.80{\pm}0.94$}
& $6.68{\pm}2.09$ \\

\texttt{nr\_ppar\_gamma}
& $10.14{\pm}1.39$
& $4.67{\pm}0.07$
& \Best{$19.38{\pm}6.42$}
& $19.30{\pm}2.85$
& $4.69{\pm}2.07$ \\

\texttt{sr\_are}
& \Best{$29.27{\pm}1.14$}
& $26.82{\pm}0.01$
& $28.81{\pm}1.70$
& $28.44{\pm}4.14$
& \NA \\

\texttt{sr\_atad5}
& $10.34{\pm}1.37$
& $7.45{\pm}0.12$
& $17.84{\pm}2.00$
& \Best{$20.91{\pm}2.64$}
& $4.65{\pm}1.15$ \\

\texttt{sr\_hse}
& $20.14{\pm}1.64$
& $9.18{\pm}0.21$
& $25.84{\pm}3.16$
& \Best{$29.91{\pm}4.19$}
& $14.47{\pm}7.13$ \\

\texttt{sr\_mmp}
& $40.13{\pm}0.20$
& $26.13{\pm}0.03$
& $41.41{\pm}4.70$
& \Best{$45.12{\pm}1.71$}
& $29.32{\pm}2.35$ \\

\texttt{sr\_p53}
& $17.60{\pm}1.34$
& $17.05{\pm}0.85$
& \Best{$30.76{\pm}7.61$}
& $26.32{\pm}2.26$
& $15.89{\pm}1.97$ \\
\bottomrule
\end{tabular}
}
\end{table}

\providecommand{\grasp}{\textsc{Grasp}}


\definecolor{graspblue}{HTML}{1F4E79}
\definecolor{grasphl}{HTML}{EAF1FB}
\definecolor{badgray}{HTML}{8A8A8A}

\newcolumntype{Y}{>{\raggedright\arraybackslash}X}

\providecommand{\rL}[1]{{\scriptsize\ttfamily #1}}
\providecommand{\gR}[1]{{\scriptsize\ttfamily\color{graspblue} #1}}
\providecommand{\wR}[1]{{\scriptsize\ttfamily\color{black} #1}}
\providecommand{\gloss}[1]{{\scriptsize\itshape\color{badgray} #1}}
\providecommand{\vtarget}[1]{\rotatebox[origin=c]{90}{\texttt{#1}}}

\begin{table}[t]
\centering
\caption{Representative top-ranked rule per method on three targets
(baselines first, \grasp{} last and highlighted with a plain-language
reading). \grasp{} invents \emph{high-level} predicates
(\rL{mol\_num\_rings}, \rL{mol\_count\_fg}), uses numeric thresholds, and
recovers chemically meaningful patterns compared to other methods.
Top
rule per method (\grasp{}: $|w_1-w_0|$; \textsc{BoostedRDN}: leaf value).}
\label{tab:rules-headtohead}
\setlength{\tabcolsep}{4pt}
\renewcommand{\arraystretch}{1.25}
\scriptsize
\begin{tabularx}{\textwidth}{@{}c l Y@{}}
\toprule
& \textbf{Method} & \textbf{Top rule} \\
\midrule

\multirow{4}{*}{\vtarget{mutagen}}
 & \textsc{BoostedRDN}  & \rL{active(A) :- ind1(A,1).} \\
 & \textsc{One-shot}    & \rL{active(A) :- lumo(A,B), B<-2.0, ind1(A,1.0)} \\
 & \textsc{NeuralLP}    & \wR{active(A) :- bond(A,V0,E0,E1), bond(V0,V1,E2,E3)} \\
 & \cellcolor{grasphl}\grasp{}
   & \cellcolor{grasphl}\gR{active(A) :- num\_atoms(A,N), N>25, lumo(A,L), L<-1.1}\newline
     \gloss{Exp: Large, low-LUMO molecules are mutagenic (size $+$ electronics).} \\
\midrule

\multirow{4}{*}{\vtarget{nr\_ar}}
 & \textsc{BoostedRDN}  & \rL{nr\_ar(A) :- fg(A,\_,ketone), fg(A,\_,allylic\_oxidation\_site).} \\
 & \textsc{One-shot}    & \rL{nr\_ar(A) :- fg(A,B,bicyclic), fg(A,C,allylic\_oxidation\_site)} \\
 & \textsc{NeuralLP}    & \wR{nr\_ar(A) :- ring\_bond(A,V0,E0,E1), smiles(V1,V0)} \\
 & \cellcolor{grasphl}\grasp{}
   & \cellcolor{grasphl}\gR{nr\_ar(A) :- mol\_ring\_count(A,2), mol\_fg\_count(A,ketone,1),
     mol\_fg\_count(A,halogen,1), mol\_fg\_count(A,aromatic\_n,1)}\newline
     \gloss{Exp: A two-ring scaffold carrying ketone, halogen and aromatic-N groups.} \\
\midrule

\multirow{4}{*}{\vtarget{sr\_mmp}}
 & \textsc{BoostedRDN}  & \rL{sr\_mmp(A) :- fg(A,\_,aromatic\_oh).} \\
 & \textsc{One-shot}    & \rL{sr\_mmp(A) :- fg(A,B,aromatic\_oh), fg(A,C,bicyclic)} \\
 & \textsc{NeuralLP}    & \wR{sr\_mmp(A) :- true}\\
 & \cellcolor{grasphl}\grasp{}
   & \cellcolor{grasphl}\gR{sr\_mmp(A) :- mol\_num\_functional\_groups(A,halogen,H), H>=3,
     mol\_num\_functional\_groups(A,ester,E), E>=2, mol\_num\_rings(A,R), R>=4}\newline
     \gloss{Exp: Many halogens $+$ esters over $\geq$4 rings drive mitochondrial toxicity.} \\
\bottomrule
\end{tabularx}
\end{table}

\textsc{grasp} learns interpretable probabilistic logic programs from relational data by combining iterative LLM proposals with functional gradient boosting. We organize our empirical evaluations aiming to answer the following questions:
\begin{enumerate}[label=\textbf{(Q\arabic*)}, leftmargin=*, itemsep=2pt]
\item[\textbf{(Q1)}] Does \textsc{grasp} learn probabilistic logic programs that achieve competitive predictive performance on held-out examples compared to symbolic and neural baselines?

\item[\textbf{(Q2)}] Does gradient-guided steering improve the quality of LLM-generated rule proposals compared to one-shot LLM generation?

\item[\textbf{(Q3)}] Do the weights assigned to rules in the ensemble reliably reflect their contribution to predictions, and does a small subset of high-weight rules suffice to explain the majority of examples?

\item[\textbf{(Q4)}] Does \textsc{grasp} recover domain knowledge in its induced rules?

\end{enumerate}






\paragraph{Datasets.} We consider $3$ relational benchmarks spanning diverse domains and relational structures for our empirical evaluation: 
\textbf{Mutagenesis} \cite{srinivasan1994mutagenesis,king1995relating}, which is a classical ILP benchmark for predicting the mutagenicity of chemical compounds from their bond and atom structure. \textbf{Cora} \cite{mccallum2000automating}, which is a popular citation network benchmark where the task is to match paper entities across duplicate records using relational features derived from authorship, venue, and title structure. \textbf{Tox21} \cite{huang2016tox21challenge}, which is a molecular toxicity prediction dataset where the task is to predict the activity of compounds against one or more toxicological targets. We consider $12$ targets from the Tox21 domain spanning both NR (Nuclear Receptor) and SR (Stress Response) categories. We represent each molecule using a relational schema encoding atomic composition, bond structure, ring membership, and functional group presence. Together, these datasets cover molecular, biomedical, and bibliographic relational domains, and vary in scale, relational depth, and predicate arity. 

\paragraph{Metrics.} We report AUC-ROC and AUC-PR across all datasets. AUC-ROC measures overall discriminative performance, while AUC-PR is particularly informative under class imbalance, which is present in several of the benchmarks. Results are averaged over 3 trials with standard deviations reported. To evaluate the calibration of the learned rule weights and the explanatory efficiency of the ensembles, we analyze insertion and deletion curves~\cite{petsiuk2018rise}. In the insertion setting, rules are cumulatively added to the ensemble in descending order of their absolute weight magnitude, while deletion starts by cumulatively deleting, starting from the most important rule.

\paragraph{Baselines.} We compare \textsc{grasp} against 3 baselines representing distinct approaches to relational learning. \textbf{NeuralLP} is a differentiable rule learning method that encodes inference as matrix operations over relation embeddings, representing the line of work that relaxes symbolic search into differentiable optimization. \textbf{BoostedRDN} is a functional gradient boosting system for statistical relational learning that uses relational regression trees as weak learners, representing the symbolic boosting approach that \textsc{grasp} directly extends. \textbf{One-shot LLM} is an ablation baseline that prompts the LLM to generate rules directly from the predicate schema and domain description without gradient-guided steering or iterative refinement, isolating the contribution of the boosting loop.
For \grasp{}, we also consider two different variants for the LLM proposal oracle - \texttt{Gemini-3.5-Flash} and \texttt{GPT-5.2-Codex}.
To reflect the interpretability requirements of probabilistic logic programs, and to ensure a fair comparison, we set the number of boosting iterations to $20$ (ensemble size) for both \grasp{} and \textsc{BoostedRDN}, ensuring that learned models remain human-inspectable and that comparisons are made under a shared interpretability constraint. For all \grasp{} experiments, we use a step size of $\eta = 1$, sample $10$ high-residual examples per prompt, and request $5$ candidate rule proposals per prompt.

\begin{table}[t]
\centering
\caption{Analysis of rule complexity and ensemble size across evaluated domains. For \textbf{Tox21 (Avg)}, metrics are averaged across all 12 experimental targets where a method successfully produced rules. The columns denote: \textbf{Total Rules} (the average number of rules generated per ensemble), \textbf{Avg Literals} (the average number of body predicates or relational depth per rule), and \textbf{Max Literals} (the maximum number of body predicates observed in any single rule for that method over the grouped tasks). All values are averaged across three trials of each method.}
\label{tab:baseline-rule-complexity}
\begin{adjustbox}{max width=\textwidth}
\begin{tabular}{llccc}
\toprule
\textbf{Domain} & \textbf{Method} & \textbf{Total Rules} & \textbf{Avg Literals} & \textbf{Max Literals} \\
\midrule
\multirow{5}{*}{Cora}
 & NeuralLP & 3.0 & 2.0 & 2 \\
 & BoostedRDN & 20.0 & 2.0 & 2 \\
 & One-shot LLM & 5.7 & 4.6 & 6 \\
 & \cellcolor{gray!15}\textbf{\textsc{grasp} (Gemini)} & \cellcolor{gray!15}11.7 & \cellcolor{gray!15}4.7 & \cellcolor{gray!15}12 \\
 & \cellcolor{gray!15}\textbf{\textsc{grasp} (Codex)} & \cellcolor{gray!15}20.0 & \cellcolor{gray!15}3.0 & \cellcolor{gray!15}8 \\
\midrule
\multirow{5}{*}{Mutagenesis}
 & NeuralLP & 33.7 & 2.7 & 3 \\
 & BoostedRDN & 20.0 & 1.6 & 2 \\
 & One-shot LLM & 16.7 & 1.8 & 6 \\
 & \cellcolor{gray!15}\textbf{\textsc{grasp} (Gemini)} & \cellcolor{gray!15}19.7 & \cellcolor{gray!15}3.4 & \cellcolor{gray!15}6 \\
 & \cellcolor{gray!15}\textbf{\textsc{grasp} (Codex)} & \cellcolor{gray!15}20.0 & \cellcolor{gray!15}1.3 & \cellcolor{gray!15}6 \\
\midrule
\multirow{5}{*}{Tox21 (Avg)}
 & NeuralLP & 115.9 & 2.0 & 3 \\
 & BoostedRDN & 20.0 & 1.8 & 2 \\
 & One-shot LLM & 7.7 & 2.3 & 3.4 \\
 & \cellcolor{gray!15}\textbf{\textsc{grasp} (Gemini)} & \cellcolor{gray!15}19.8 & \cellcolor{gray!15}3.3 & \cellcolor{gray!15}6.1 \\
 & \cellcolor{gray!15}\textbf{\textsc{grasp} (Codex)} & \cellcolor{gray!15}19.4 & \cellcolor{gray!15}1.2 & \cellcolor{gray!15}2.5 \\
\bottomrule
\end{tabular}
\end{adjustbox}
\end{table}


\paragraph{Results.}
\begin{enumerate}[label=\textbf{(Q\arabic*)}, leftmargin=*, itemsep=2pt]
    \item[\textbf{(Q1)}] Tables~\ref{tab:aucroc-all-domains} and~\ref{tab:aucpr-all-domains} report AUC-ROC and AUC-PR across all benchmarks. \textsc{grasp} achieves the best result on the majority of datasets, with \textsc{grasp}-Cdx (\texttt{GPT-5.2 Codex}) and \textsc{grasp}-Gem (\texttt{Gemini-3.5-flash}) jointly obtaining the highest AUC-ROC on $10$ and highest AUC-PR on $12$ of the $14$ datasets. Notably, \textsc{grasp} outperforms NeuralLP across nearly all domains, and matches or exceeds BoostedRDN on most Tox21 nuclear receptor and citation targets. Overall, these results confirm that \textsc{grasp} learns accurate probabilistic logic programs, answering {\bf (Q1)} affirmatively.
    \item[\textbf{(Q2)}] Tables~\ref{tab:aucroc-all-domains} and~\ref{tab:aucpr-all-domains} also report results for \textsc{grasp} and the one-shot LLM baseline across all four benchmarks. The one-shot baseline is consistently the weakest method across both metrics, and sometimes fails to generate any valid rules. It also shows substantially higher variance, with standard deviations reaching 21.6 AUC-ROC points on Cora and 17.8 AUC-PR points, compared to below 3 points for \textsc{grasp} variants on the same dataset. This confirms that LLM rule generation without gradient-guided steering is unreliable, and that the pseudo-residual feedback signal is the key driver of \textsc{grasp}'s performance.

\begin{figure}
    \centering
    \includegraphics[width=0.85\linewidth]{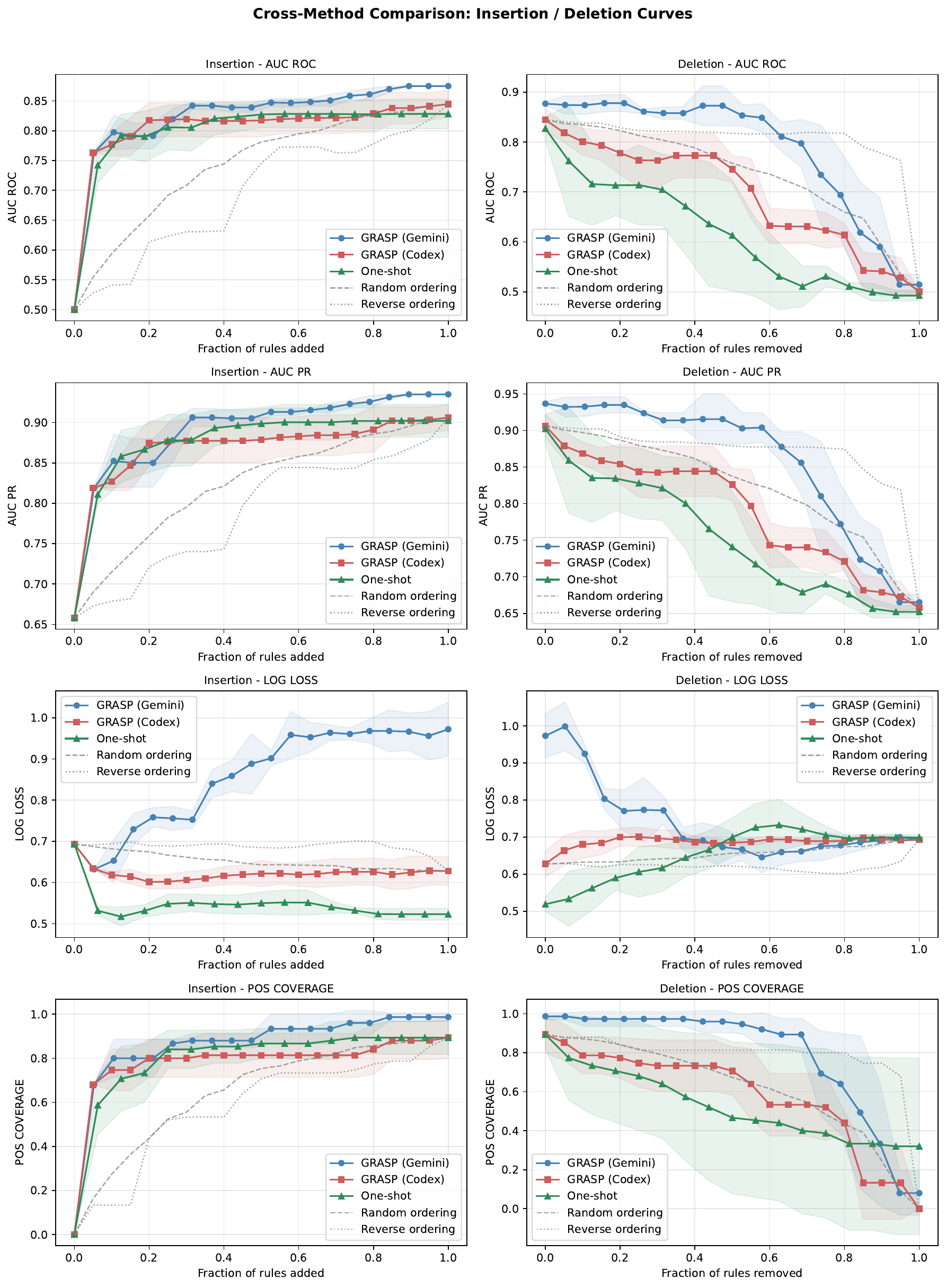}
    \caption{Evaluation of rule importance and ensemble efficiency via insertion (top) and deletion (bottom) curves, ordered by absolute discriminative weight. Gray dashed, and dotted lines denote random and reverse-ordering ablations for \textsc{grasp} (Codex). \textsc{grasp} models exhibit steep insertion asymptotes and rapid deletion degradation (outperforming baselines), confirming accurate weight calibration and demonstrating that a highly compact subset of rules captures the majority of predictive variance. Initial deletion robustness in \textsc{grasp} (Codex) reflects semantic redundancy among top-weighted rules. 
}
    \label{fig:insertion}
\end{figure}
    \item[\textbf{(Q3)}] Figure~\ref{fig:insertion} shows the insertion and deletion curves for the mutagenesis dataset.  We observe a steep, monotonic improvement in insertion curves for both AUC-ROC and AUC-PR, accompanied by deletion curves showing an immediate and precipitous degradation in predictive performance. 
    
    While the insertion curves confirm that the gradient-guided LLMs accurately assign the highest weights to the most discriminative rules, the deletion curves reveal a distinct characteristic of LLM-proposed ensembles: high semantic redundancy. For \textsc{grasp} (Codex), the deletion trajectory remains initially higher than the random baseline. This indicates that the LLM generates multiple, highly weighted structural variations of the most important relational patterns. As a result, ablating the top-ranked rule does not immediately degrade predictive performance or positive coverage (Figure~\ref{fig:insertion}, bottom right), as functionally redundant rules remain in the ensemble to score those examples. While this redundancy provides ensembling robustness, it suggests that future iterations of \textsc{grasp} could benefit from an explicit coverage-overlap penalty during the boosting step to force the LLM to explore more diverse hypothesis spaces.

    \par Contextualizing these curves with the ensemble complexity metrics (Table \ref{tab:baseline-rule-complexity}), where \textsc{grasp} induces an average of approximately 12 to 20 rules per task, peak performance is consistently recovered within the first 20\% of the ensemble. This demonstrates that a highly compact subset of 2 to 4 high-weight rules is sufficient to explain the majority of the relational variance. These findings confirm that \textsc{grasp} learns succinct and highly interpretable probabilistic logic programs.
    
    \item[\textbf{(Q4)}] To evaluate whether \textsc{grasp} induces rules that capture meaningful relational structure, we examine the top-ranked rule per target shown in Table~\ref{tab:rules-headtohead} and assess their consistency with established domain knowledge. For mutagenesis, the rule identifies large molecules with low electron affinity as mutagenic, consistent with established structure-activity relationships~\cite{compadre1990}. For androgen receptor antagonism (\texttt{nr\_ar}), the two-ring scaffold carrying a ketone, halogen, and aromatic nitrogen is consistent with structural features associated with this toxicity endpoint in the Tox21 chemical collection~\cite{lynch2017,kleinstreuer2017}. For mitochondrial membrane potential disruption (\texttt{sr\_mmp}), the combination of multiple halogens, ester groups, and a large ring count is associated with increased lipophilicity and membrane permeability, properties strongly linked to mitochondrial toxicity~\cite{attene2014}. Taken together, these results suggest that \textsc{grasp} induces rules consistent with established domain knowledge, producing interpretable outputs grounded in known biochemical relationships.

\end{enumerate}

\section{Conclusion}
We introduced \textsc{grasp}, a neurosymbolic framework that learns interpretable probabilistic logic programs from relational data by combining iterative LLM proposals with functional gradient boosting. Experiments across citation matching, mutagenesis, and molecular toxicity domains show that \textsc{grasp} outperforms classical symbolic and neural baselines under a shared interpretability constraint, and that gradient-guided steering is essential for reliable rule synthesis. Analysis of the induced rules confirms that \textsc{grasp} recovers relational patterns consistent with established domain knowledge.

Several directions remain open for future work. First, \textsc{grasp} currently operates over symbolic predicates defined over noisy structured relational data. Extending the framework to support neural predicates would allow \textsc{grasp} to operate over unstructured inputs such as images and text. Second, the interpretable rule ensembles produced by \textsc{grasp} open a natural avenue for human-allied learning, where domain experts can inspect and refine induced rules, aggregation predicates, and entity representations. Finally, \textsc{grasp} currently treats the LLM as a black-box proposer; fine-tuning the LLM  could improve the quality and validity of proposed rules. 

\section*{Acknlowedgements}
\textit{Sahil Sidheekh} and \textit{Sriraam Natarajan} gratefully acknowledge support from the AFOSR award FA9550-23-1-0239, the ARO award W911NF2010224, and the DARPA Assured Neuro Symbolic Learning and Reasoning award HR001122S0039.\textit{ Saurabh Mathur} and \textit{Kristian Kersting} acknowledge support from the Cluster of Excellence ``Reasonable AI,'' funded by the German Research Foundation (DFG) under Germany's Excellence Strategy, EXC-3057. \textit{Bhavan Vasu} and \textit{Prasad Tadepalli} acknowledge support from the Army Research Office under grant W911NF2210251. This research was conducted in part while Sahil Sidheekh was a summer research intern at AT\&T, funded through Tech Mahindra (Americas) Inc. The authors also thank the AT\&T CDO team for providing the H100 GPU infrastructure and API credits that supported these experiments.


%
%
\bibliographystyle{splncs04}
\bibliography{IJCLR/refs}

\end{document}